# End-to-end autonomous scientific discovery on a real optical platform

*Shuxing Yang[1,2,3], Fujia Chen[1,2,3], Rui Zhao[1,2,3], Junyao Wu[1,2,3], Yize Wang[1,2,3],*

*Haiyao Luo[1,2,3], Ning Han[4], Qiaolu Chen[1,2,5], Yuze Hu[1], Wenhao Li[1,2,3], Mingzhu Li[6],*

*Hongsheng Chen[*,1,2,3], and Yihao Yang[*,1,2,3]*

[1] State Key Laboratory of Extreme Photonics and Instrumentation, College of Information Science and Electronic Engineering, ZJU-Hangzhou Global Scientific and Technological Innovation Center, The Electromagnetics Academy at Zhejiang University, Zhejiang University, Hangzhou 310027, China

[2] Key Laboratory of Advanced Micro/Nano Electronic Devices & Smart Systems of Zhejiang, Jinhua Institute of Zhejiang University, Zhejiang University, Jinhua 321099, China

[3] Shaoxing Institute of Zhejiang University, Zhejiang University, Shaoxing 312000, China

[4] College of Optical and Electronic Technology, China Jiliang University, Hangzhou 310018, China

[5] Laboratory of Wave Engineering, Ecole Polytechnique Fédérale de Lausanne (EPFL), Lausanne, Switzerland

[6] School of Information and Electrical Engineering, Hangzhou City University, Hangzhou 310015, China

Correspondence should be addressed to Yihao Yang (yangyihao@zju.edu.cn) and Hongsheng Chen (hansomchen@zju.edu.cn).

# ABSTRACT

Scientific research has long been human-led, driving new knowledge and transformative technologies through the continual revision of questions, methods and claims as evidence accumulates. Although large language model (LLM)-based agents are beginning to move beyond assisting predefined research workflows, none has yet demonstrated end-to-end autonomous discovery in a real physical system that produces a nontrivial result supported by experimental evidence. Here we introduce Qiushi Discovery Engine, an LLM-based agentic system for end-to-end autonomous scientific discovery on a real optical platform. Qiushi Engine combines nonlinear research phases, Meta-Trace memory and a dual-layer architecture to maintain adaptive and stable research trajectories across long-horizon investigations involving thousands of LLM-mediated reasoning, measurement and revision actions. It autonomously reproduces a published transmission-matrix experiment on a non-original platform and converts an abstract coherence-order theory into experimental observables, providing, to our knowledge, the first observation of this class of coherence-order structure. More importantly, in an open-ended study involving 145.9 million tokens, 3,242 LLM calls, 1,242 tool calls, 163 research notes and 44 scripts, Qiushi Engine proposes and experimentally validates optical bilinear interaction, a physical mechanism structurally analogous to a core operation in Transformer attention. This AI-discovered mechanism suggests a route towards high-speed, energy-efficient optical hardware for pairwise computation. To our knowledge, this is the first demonstration of an AI agentic system autonomously identifying and experimentally validating a nontrivial, previously unreported physical mechanism, marking a milestone for research-level autonomous agents. More broadly, it signals an emerging paradigm shift from AI-assisted research to AI-led scientific discovery.

**Introduction**

Scientific discovery has long been a human-led process and remains a primary source of new knowledge and transformative technologies[1]. Yet discovery rarely emerges from a short sequence of well-defined operations. It is produced through a complex, long-horizon research process that can involve hundreds or thousands of interdependent reasoning and action steps across interpretation, modelling, implementation, measurement and revision[2,3]. This process is not fixed in advance. Researchers do not simply solve a predefined problem and stop; they move between ideas, methods, experiments and claims, often returning to earlier stages when new evidence changes what should be tested, how it should be tested or what can be claimed. Scientific research is also grounded in physical reality. Theories and computational plans must be translated into executable procedures, confronted with imperfect instruments and noisy measurements, and revised when the measured world deviates from the idealized one. These features have made scientific discovery one of the most difficult human intellectual activities to automate end to end[3].

Recent large language model (LLM)-based systems have advanced autonomous scientific research along several directions[4–13]. At one level, LLMs serve as scientific assistants for literature analysis, hypothesis generation, protocol planning, coding, data analysis and writing[6,8,11,12]. At another, LLM agents are connected to expert tools, simulators, robotic platforms or scientific instruments, enabling bounded workflows in chemistry, materials science, biology and laboratory automation[4,7,14–17]. A third line of work organizes multiple LLM agents into virtual laboratories or end-to-end digital research pipelines, producing hypotheses, code, experiments, analyses and manuscripts with increasing autonomy[5,9,10,13]. These studies show that LLM can coordinate substantial segments of the research process.

However, most existing systems remain constrained in three respects. They are often workflow-bound[4–17], because the route, objective or evaluation criterion is specified in advance; environment-bound, because many operate primarily in digital, symbolic or highly controlled settings; and horizon-bound, because the tasks do not require sustained reorganization of a research trajectory over hundreds or thousands of model calls under continuing feedback from the real world. As a result, existing systems have not yet demonstrated end-to-end autonomous scientific discovery in the stronger sense required here: starting from an open-ended scientific theme, interacting with a real physical system, and delivering a nontrivial new discovery supported by experimental evidence.

Here we introduce Qiushi Discovery Engine, an LLM-based agentic system for long-horizon end-to-end autonomous scientific discovery on a real optical platform. It sustains adaptive and stable research trajectories across thousands of LLM-mediated reasoning, tool-use and revision steps while interacting with physical experiments. Adaptability arises from a nonlinear Explore-Execute-Express flow that is decoupled from four agent roles, allowing the system to reorganize its trajectory as evidence accumulates. Stability is provided by Meta-Trace and a dual-layer architecture, which preserve structured research state without overwhelming the main reasoning context.

The system is coupled to a free-space optical platform. Free-space optics provides a stringent testbed for autonomous scientific discovery: it is central to imaging, sensing and optical information processing[18–21], yet requires abstract wave theory, high-dimensional field control, hardware calibration and direct measurement to be integrated in a single physical setting. We evaluate Qiushi Engine through three progressively more demanding studies. First, it transfers a published transmission-matrix experiment[22] to a non-original optical platform. Second, it converts an abstract coherence-order theory[23–26] into experimentally testable transport observables and validates the predicted ordering relation using measured optical operators, providing, to our knowledge, the first experimental validation of this class of coherence-order structure in optics. Both studies are completed within hours, whereas comparable work would typically require trained graduate researchers weeks to months of sustained effort.

Most importantly, Qiushi Engine moves beyond reproduction and theory-to-experiment translation into open-ended exploration. Starting from a broad theme at the intersection of optical computing and artificial intelligence[19,20,27–32], the system conducts a 206-step autonomous study over 1,288.1 min, using 145.9 million tokens, 3,242 LLM calls and 1,242 tool invocations, while producing 163 research notes and 44 scripts. Through this trajectory, it identifies optical bilinear interaction as a previously unreported physical mechanism in which coherent scattering and square-law detection generate pairwise optical features. This mechanism is structurally analogous to the bilinear compatibility calculation between query and key representations in Transformer attention[33,34], and suggests a route towards high-speed, energy-efficient optical hardware for pairwise computation. To our knowledge, this is the first demonstration of an AI agentic system autonomously proposing and experimentally validating a nontrivial physical mechanism in a real experimental environment, marking a milestone towards AI-led scientific discovery.

**Results**

Qiushi Engine is a dual-layer multi-agent system for autonomous scientific discovery in a physically grounded setting. The name "Qiushi", meaning "seeking truth", reflects the system's evidence-grounded research principle. Real scientific research requires planning, theory-to-method construction, execution, evaluation, memory, retrieval, exploration and verification to operate together over extended trajectories. The central architectural challenge is to coordinate these functions without allowing auxiliary information, tool traces or experimental artefacts to destabilize the main scientific reasoning process. Qiushi Engine addresses this challenge by separating agents that carry the main research trajectory from agents that provide context-isolated support functions (Fig. 1a). The core research agent system maintains the evolving research direction, while the support research agent system supplies memory, retrieval, auxiliary exploration and evidence verification through structured interfaces. Both operate on a shared infrastructure layer comprising a physical interface to the experiment and a digital execution environment for files, code, data and simulation. This separation preserves a stable, evidence-constrained research state while allowing the system to interact continuously with literature, code, data, tools and real physical experiments.

The core research agent system contains four role-specialized agents: a Lead Investigator for global planning, hypothesis formation and trajectory control; a Method Builder for theory-to-method translation, algorithm design and manuscript construction; an Experimentalist for simulation, code execution, physical measurement and data analysis; and a Critical Reviewer for adversarial evaluation of evidence, claims and limitations. These agents represent complementary modes of scientific reasoning[4,10] rather than stages of a fixed pipeline, enabling nonlinear movement between planning, construction, execution and critique as evidence accumulates. The support research agent system contains context-isolated sub-agents for history review, retrieval, hypothesis exploration, trajectory tracking and evidence verification. Core agents communicate with support sub-agents through structured requests and curated returns: they request specific information, checks or explorations, and receive compressed, task-relevant outputs rather than raw search traces, tool logs or deliberation histories. This interface allows Qiushi Engine to use memory, retrieval, exploration and independent verification without overwhelming the active research context, preserving trajectory coherence while enabling revision when experiments fail or evidence reshapes the claim boundary.

We couple Qiushi Engine to a real free-space optical platform through a standardized physical interface (Fig. 1b). The platform provides a high-dimensional space for optical control and measurement: more than two million 10-bit-addressed SLM pixels shape the incident field, corresponding to an optical-control space on the order of $2^{20,000,000}$ configurations, while camera-based detection records tens of millions of output pixels after scattering. A diffuser introduces strong mode mixing, producing distributed speckle patterns and a nonlocal input–output relation that is difficult to predict from local optical intuition alone[18,21]. Additional details of the experimental setup are provided in Methods. This combination of high dimensionality, strong mode mixing and direct physical measurement makes the platform a stringent testbed for physically grounded autonomous scientific research.

Qiushi Engine is designed to be adaptive rather than pipeline-bound. Scientific research is organized into three nonlinear phases—Explore, Execute and Express—but these phases are deliberately decoupled from the four role-specialized agents. Explore includes literature interpretation, hypothesis generation, theory mapping and observable design; Execute includes coding, simulation, physical experiment and data analysis; and Express includes figure construction, manuscript writing, evidence synthesis and critical review. Because any core agent can operate in any phase (Fig. 1c; Methods), each Agent Step can occupy one of 12 role-phase configurations (Supplementary Note 1). An *n*-step investigation therefore spans, in principle, up to $12^n$ possible role-phase trajectories. Qiushi Engine does not enumerate this combinatorial space; it selects and revises a path through it according to the evolving state of evidence. A failed measurement may return the system from execution to observable design, a review step may trigger new simulation or experiment, and a manuscript draft may expose an unsupported claim and force further analysis. This role-phase decoupling is the basis of the system's adaptivity: the research trajectory is not predefined, but reorganized as evidence changes what should be tested, how it should be tested and what can be claimed. Beyond single-session adaptation, the workflow includes an experience-consolidation process that turns completed studies into reusable research experience, enabling strategies and claim–evidence patterns to accumulate across investigations (Supplementary Note 1).

The openness that enables adaptivity also creates a stability problem. Over long runs, the system must preserve the scientific state across thousands of LLM calls, tool invocations, files, figures, experimental logs and intermediate judgments. Qiushi Engine addresses this challenge through Meta-Trace together with the dual-layer architecture. Meta-Trace is not a passive transcript or conversational summary. At each Agent Step boundary, the acting agent distils the step into a structured unit of scientific know-how: what was attempted, what was found, which evidence supports the current state, what limitations remain, what artefacts were produced and what the next agent should do (Methods). In parallel, the system maintains a set of auditable research records, including scripts, notes, figures, experimental parameters, measurement outputs, tool calls and reports. These mechanisms expose the information needed for continued reasoning while preventing raw traces from flooding the active context (Supplementary Note 1).

Information flow in Qiushi Engine is organized across two levels: step-to-step handoff and within-step action (Fig. 1d; Methods). At the boundary between Agent Steps, the system updates the context exposed to the next acting agent, including the system prompt, short-term memory, condensed Meta-Trace memory and progressively disclosed knowledge and skills. This handoff defines the next agent's objective, role, accessible history and relevant operational knowledge, allowing the research trajectory to continue without exposing the agent to the full raw record. Within each Agent Step, the acting agent uses this updated context to perform local research actions. It can request history review, retrieval, auxiliary exploration, trajectory analysis or evidence verification from support sub-agents, and it can act through the infrastructure layer by reading and writing files, executing code, running simulations, producing figures, analysing data or invoking calibrated routines on the physical optical platform. The resulting tool outputs, experimental logs and verifier responses are curated into task-specific evidence, decisions and artefacts rather than appended directly as uncontrolled context. The step ends with a new Meta-Trace entry that records the current state, evidence, limitations and next handoff. This two-level information flow allows bounded LLM calls to support long-horizon, coherent and auditable research trajectories.

We first ask whether Qiushi Engine can reproduce a published transmission-matrix experiment on a non-original optical platform. The target study, published in Physical Review Letters (PRL), recovers a monochromatic transmission matrix interferometrically and uses it to focus light through a scattering medium[22]. Because the local free-space platform differs from the original setup in optical layout, modulation interface, reference geometry, detection and calibration, Qiushi Engine must translate the published measurement logic into a new experimental setting and determine which claims remain supported under local constraints (Fig. 2a).

Starting only from the PRL paper and a basic platform description, Qiushi Engine extracts the central objective—measuring the complex input–output response of the scattering medium for phase-conjugate focusing—and maps it onto the local hardware. It repairs the software–hardware interface, designs calibrated phase-stepped measurements, begins with pilot acquisitions and then scales to the full transmission-matrix experiment (Fig. 2b). As data accumulate, the system moves between code, measurement and analysis in a manner closer to a mature experimental researcher than to a fixed workflow: it tests partial implementations, checks whether the recovered matrix produces the expected focusing behaviour[18,35], and revises the next action according to the evidence. The transition from Agent Step 17 to Step 18 illustrates this behaviour (Fig. 2c): after obtaining the main focusing effect, the system asks whether the evidence also supports stronger image- or pattern-reconstruction claims. The Critical Reviewer judges that it does not, prompting a targeted follow-up experiment whose negative evidence bounds the final conclusion (Supplementary Note 2).

Over 50 Agent Steps and 366.4 min, Qiushi Engine uses 27.6 million tokens, 482 LLM calls and 439 tool calls to complete the reproduction study. The primary 256 × 256 transmission-matrix acquisition comprises 1,025 calibrated measurements and supports phase-conjugate focusing (Fig. 2d,e). It further identifies that focusing enhancement increases with the number of controlled input modes[35] (Fig. 2f). In a follow-up experiment, the system screens annular reference-field geometries and improves the best-case enhancement from 25.59 to 46.1 (Fig. 2g; Methods; Supplementary Note 2). These results show that Qiushi Engine can transfer a published optical protocol to a different physical platform, bound the supported claim and improve the implementation under local experimental constraints.

We next ask whether Qiushi Engine can go beyond experimental reproduction and design a validation study for an abstract theory that has not yet been experimentally tested on a real optical platform. The target is the recent theory of majorization order in wave coherence, which predicts that coherence ordering is manifested in transport measurements under unitary control[26,36–38]: a less coherent spectrum in the majorization order[23–25] should have a more restricted achievable transport-response range than a more coherent spectrum. The challenge is that this prediction is not a directly visible camera image or a simple intensity observable, but a relation between achievable response intervals that must be translated into platform-specific measurements through a theory-to-experiment trajectory (Fig. 3a,b).

Starting from the theory paper, the platform description and prior transmission-matrix experience, Qiushi Engine constructs the theory-experiment interface needed for this validation. It recognizes that raw camera intensity in the self-referenced architecture contains reference background and interference terms, and therefore cannot directly serve as the transport observable in the theory. It also recognizes that the platform directly launches coherent fields, so mixed-state coherence spectra[25,39] must be implemented through deterministic weighted reconstruction. On this basis, the system reformulates the validation around transmission-matrix-derived transport operators: it measures a transmission matrix, constructs effective operators from the measured data and compares the resulting transport intervals for selected coherence spectra (Methods). The early handoff from the first to the second Agent Step illustrates this translation, with an initial validation plan converted into concrete measurement and analysis operations accurately on the optical platform (Fig. 3c).

Over a 38-step study, Qiushi Engine measures a self-referenced 16-port transmission matrix and constructs a family of effective transport operators from the measured data (Fig. 3d–f). For all tested comparable pairs, the transport-response interval of the less coherent spectrum remains nested within that of the more coherent spectrum across the measured readout systems. In contrast, tested incomparable pairs do not show a universal nesting order; instead, they exhibit partial interval overlap in at least one readout system, including the benchmark incomparable cases from the target theory (Supplementary Note 3). To our knowledge, this provides the first experimental validation of this transport prediction on a real optical platform. The study also demonstrates a distinct capability of the system: Qiushi Engine does not merely reproduce procedures already present in the literature, but converts an abstract theoretical result without a ready-made experimental protocol into a physically executable validation study. The complete process is finished within hours—over 38 Agent Steps and 175.8 min of autonomous operation, with 22.06 million tokens, 337 LLM calls and 182 tool calls—whereas a comparable theory-to-experiment translation would typically require trained graduate researchers weeks to months of sustained work.

Having shown that Qiushi Engine can transfer a published experiment to a non-original platform and convert an abstract theory into a real-platform validation study, we next test whether it can conduct open-ended discovery. We choose the intersection of optical computing and artificial intelligence because it is intrinsically cross-disciplinary: progress requires concepts from optical engineering, wave physics, machine-learning architectures and computational primitives. Such a setting is well suited to autonomous exploration because the valuable research direction is not specified in advance and may emerge only by connecting ideas across fields. The initial prompt therefore specifies only a broad theme—optical computing for artificial intelligence—without prescribing the target phenomenon, observable or experimental protocol.

Given this prompt, Qiushi Engine sustains a 206-step trajectory over 1,288.1 min, using 145.9 million tokens, 3,242 LLM calls and 1,242 tool invocations (Fig. 4a,b). It first generates and develops four candidate directions (Fig. 4c): deterministic physical token embedding in complex scattering media (Steps 1-38), bilinear interaction engines for compact optical pairwise computation (Steps 39-71), family-controlled physical interaction geometry (Steps 72-98), and scale-dependent specialization and degeneracy boundaries in family-programmable scattering engines (Steps 99-162). After this autonomous exploration stage, the second direction is selected for further refinement (Steps 163-206).

The origin of this idea is captured by the Meta-Trace at Agent Step 39 (Fig. 4d). After exploring the scattering medium as a deterministic single-token embedding engine, Qiushi Engine judges that this framing remains close to established random-feature paradigms[40] and searches for a more distinctive optical-computing mechanism. The new idea is derived from three platform properties: coherent superposition of independently encoded optical fields, high-dimensional mixing by the scattering medium and square-law detection by the camera. Combining these ingredients with the prior single-token embedding experience, the system reasons that two optical inputs should generate a measurable cross-term, allowing the scattering platform to support a token-token interaction rather than only two separate token embeddings. Thus, the optical bilinear-interaction idea is not an arbitrary suggestion, but a mechanism inferred from the physical properties of the platform and the system's accumulated research trajectory.

The resulting mechanism is an optical bilinear interaction, structurally related to a core operation in Transformer attention[33,34] (Fig. 4e,f). In attention, two tokens are projected into query and key representations, and their compatibility is computed through a bilinear operation between these representations. Qiushi Engine proposes an optical analogue of this interaction: two independently encoded optical fields are coherently superposed with controlled relative phase, scattered by a fixed medium and measured through square-law detection. Four-phase interferometric demodulation with blank subtraction isolates a complex channel-wise bilinear term at each detector channel. The result is a measurable complex interaction field that depends on the joint structure of the two inputs rather than on either input alone, providing a physical route to pairwise relational computation in optics[28–30] (Methods; Supplementary Note 4).

Qiushi Engine validates this mechanism through two physical experiments on the self-referenced scattering platform: a four-token XOR experiment and an eight-token semantic benchmark. The XOR task is especially informative because XOR-type relations cannot be separated by a purely linear map of individual inputs; they require an interaction term or nonlinear processing. Here, the optical bilinear interaction supplies such a pairwise physical feature, allowing the measured field to resolve both pair identity and XOR parity. In the semantic benchmark, the extracted Complex-B fields form distinct channel distributions for different ordered pairs (Fig. 4g,h). Under matched linear evaluation, Complex-B simultaneously preserves pair identity, same-category relation and category-pair structure, whereas token concatenation and an intensity-only digital bilinear baseline each fail on one of these axes (Fig. 4i). These results show that the measured Complex-B field is not merely an intensity pattern, but a physically generated pair-dependent representation (Methods).

This discovery is significant for both optical computing and autonomous scientific discovery. For optical computing, the identified optical bilinear interaction provides an experimentally validated primitive for pairwise relational computation[19,20,28–30], a central operation in attention-based AI[33,34]. It suggests a route towards high-speed and energy-efficient optical hardware for Transformer-like computation[31], in which part of the relational operation is generated by coherent wave physics rather than electronic post-processing alone. For autonomous scientific discovery, the result is unexpected and exceeds a conventional reproduction or validation task: from an open-ended cross-disciplinary prompt, Qiushi Engine identifies a nontrivial research direction, formulates a physical mechanism, designs the measurement protocol and validates it experimentally. To our knowledge, this is the first demonstration of an AI agentic system autonomously proposing and experimentally validating a nontrivial, previously unreported physical mechanism in a real experimental environment.

## Conclusions

Qiushi Engine demonstrates autonomous, experimentally grounded research at a scale and depth that cannot be reduced to short task execution or workflow automation. Across the studies reported here, the system sustains hundreds of Agent Steps, thousands of LLM calls and tool interactions, and up to approximately 150 million tokens of long-horizon reasoning while remaining coupled to a real optical platform. The significance of this operating regime is not scale alone, but the ability to preserve, revise and complete a research trajectory under physical constraint. The three studies form an increasing sequence of research autonomy: transferring a published protocol to a different platform, turning an abstract coherence-order prediction into a measured validation, and finally formulating and testing a new optical bilinear-interaction mechanism from an open-ended prompt. These results show that an AI agentic system can move beyond assisting predefined scientific tasks towards autonomously producing experimentally grounded knowledge.

The free-space optical platform used here also defines a broad physical discovery space in its own right. Beyond the specific studies reported in this work, such a platform can support autonomous exploration in optical computing, imaging, sensing, wavefront control, complex-medium transport and high-dimensional light-field manipulation, and can be further extended towards quantum optics and nonlinear optics. Thus, Qiushi Engine is not confined to a single optical demonstration, but provides a route for sustained autonomous investigation across a wide range of experimentally grounded optical problems.

Although demonstrated here in optics, this capability is not specific to optics. Many scientifically important domains require the same class of reasoning: abstract theory must be connected to imperfect instruments, noisy measurements, numerical modelling, evolving hypotheses and evidence-bounded claims. By showing that an agentic system can maintain such a trajectory in a real physical environment, Qiushi Engine establishes a concrete step towards AI-led scientific discovery. More broadly, it suggests a future research paradigm in which autonomous systems are not only tools for analysis or automation, but active participants in the formulation, execution and validation of scientific discoveries across materials science, quantum-device research, chemistry, biology and other research domains.

# References

1. Wang, H., Fu, T., Du, Y., Gao, W., Huang, K., Liu, Z., Chandak, P., Liu, S., Van Katwyk, P., Deac, A., Anandkumar, A., Bergen, K., Gomes, C. P., Ho, S., Kohli, P., Lasenby, J., Leskovec, J., Liu, T.-Y., Manrai, A., Marks, D., Ramsundar, B., Song, L., Sun, J., Tang, J., Veličković, P., Welling, M., Zhang, L., Coley, C. W., Bengio, Y. & Zitnik, M. Scientific discovery in the age of artificial intelligence. *Nature* **620**, 47–60 (2023).

2. Langley, P., Simon, H. A., Bradshaw, G. L. & Zytkow, J. M. *Scientific Discovery: Computational Explorations of the Creative Process*. (MIT Press, 1987). doi:10.7551/mitpress/6090.001.0001.

3. Waltz, D. & Buchanan, B. G. Automating Science. *Science* **324**, 43–44 (2009).

4. Boiko, D. A., MacKnight, R., Kline, B. & Gomes, G. Autonomous chemical research with large language models. *Nature* **624**, 570–578 (2023).

5. Lu, C., Lu, C., Lange, R. T., Foerster, J., Clune, J. & Ha, D. The AI Scientist: Towards Fully Automated Open-Ended Scientific Discovery. Preprint at https://doi.org/10.48550/arXiv.2408.06292 (2024).

6. Gao, S., Fang, A., Huang, Y., Giunchiglia, V., Noori, A., Schwarz, J. R., Ektefaie, Y., Kondic, J. & Zitnik, M. Empowering biomedical discovery with AI agents. *Cell* **187**, 6125–6151 (2024).

7. Bran, A. M., Cox, S., Schilter, O., Baldassari, C., White, A. D. & Schwaller, P. Augmenting large language models with chemistry tools. *Nat. Mach. Intell.* **6**, 525–535 (2024).

8. Romera-Paredes, B., Barekatain, M., Novikov, A., Balog, M., Kumar, M. P., Dupont, E., Ruiz, F. J. R., Ellenberg, J. S., Wang, P., Fawzi, O., Kohli, P. & Fawzi, A. Mathematical discoveries from program search with large language models. *Nature* **625**, 468–475 (2024).

9. Gottweis, J., Weng, W.-H., Daryin, A., Tu, T., Palepu, A., Sirkovic, P., Myaskovsky, A., Weissenberger, F., Rong, K., Tanno, R., Saab, K., Popovici, D., Blum, J., Zhang, F., Chou, K., Hassidim, A., Gokturk, B., Vahdat, A., Kohli, P., Matias, Y., Carroll, A., Kulkarni, K., Tomasev, N., Guan, Y., Dhillon, V., Vaishnav, E. D., Lee, B., Costa, T. R. D., Penadés, J. R., Peltz, G., Xu, Y., Pawlosky, A., Karthikesalingam, A. & Natarajan, V. Towards an AI co-scientist. Preprint at https://doi.org/10.48550/arXiv.2502.18864 (2025).

10. Swanson, K., Wu, W., Bulaong, N. L., Pak, J. E. & Zou, J. The Virtual Lab of AI agents designs new SARS-CoV-2 nanobodies. *Nature* **646**, 716–723 (2025).

11. Asai, A., He, J., Shao, R., Shi, W., Singh, A., Chang, J. C., Lo, K., Soldaini, L., Feldman, S., D'Arcy, M., Wadden, D., Latzke, M., Sparks, J., Hwang, J. D., Kishore, V., Tian, M., Ji, P., Liu, S., Tong, H., Wu, B., Xiong, Y., Zettlemoyer, L., Neubig, G., Weld, D. S., Downey, D., Yih, W., Koh, P. W. & Hajishirzi, H. Synthesizing scientific literature with retrieval-augmented language models. *Nature* **650**, 857–863 (2026).

12. Zhao, W., Wu, C., Fan, Y., Qiu, P., Zhang, X., Sun, Y., Zhou, X., Zhang, S., Peng, Y., Wang, Y., Sun, X., Zhang, Y., Yu, Y., Sun, K. & Xie, W. An agentic system for rare disease diagnosis with traceable reasoning. *Nature* **651**, 775–784 (2026).

13. Lu, C., Lu, C., Lange, R. T., Yamada, Y., Hu, S., Foerster, J., Ha, D. & Clune, J. Towards end-to-end automation of AI research. *Nature* **651**, 914–919 (2026).

14. Burger, B., Maffettone, P. M., Gusev, V. V., Aitchison, C. M., Bai, Y., Wang, X., Li, X., Alston, B. M., Li, B., Clowes, R., Rankin, N., Harris, B., Sprick, R. S. & Cooper, A. I. A mobile robotic chemist. *Nature* **583**, 237–241 (2020).

15. Szymanski, N. J., Rendy, B., Fei, Y., Kumar, R. E., He, T., Milsted, D., McDermott, M. J., Gallant, M., Cubuk, E. D., Merchant, A., Kim, H., Jain, A., Bartel, C. J., Persson, K., Zeng, Y. & Ceder, G. An autonomous laboratory for the accelerated synthesis of inorganic materials. *Nature* **624**, 86–91 (2023).

16. Dai, T., Vijayakrishnan, S., Szczypiński, F. T., Ayme, J.-F., Simaei, E., Fellowes, T., Clowes, R., Kotopanov, L., Shields, C. E., Zhou, Z., Ward, J. W. & Cooper, A. I. Autonomous mobile robots for exploratory synthetic chemistry. *Nature* **635**, 890–897 (2024).

17. Tom, G., Schmid, S. P., Baird, S. G., Cao, Y., Darvish, K., Hao, H., Lo, S., Pablo-García, S., Rajaonson, E. M., Skreta, M., Yoshikawa, N., Corapi, S., Akkoc, G. D., Strieth-Kalthoff, F., Seifrid, M. & Aspuru-Guzik, A. Self-Driving Laboratories for Chemistry and Materials Science. *Chem. Rev.* **124**, 9633–9732 (2024).

18. Mosk, A. P., Lagendijk, A., Lerosey, G. & Fink, M. Controlling waves in space and time for imaging and focusing in complex media. *Nat. Photonics* **6**, 283–292 (2012).

19. Wetzstein, G., Ozcan, A., Gigan, S., Fan, S., Englund, D., Soljačić, M., Denz, C., Miller, D. A. B. & Psaltis, D. Inference in artificial intelligence with deep optics and photonics. *Nature* **588**, 39–47 (2020).

20. Shastri, B. J., Tait, A. N., Ferreira de Lima, T., Pernice, W. H. P., Bhaskaran, H., Wright, C. D. & Prucnal, P. R. Photonics for artificial intelligence and neuromorphic computing. *Nat. Photonics* **15**, 102–114 (2021).

21. Cao, H., Mosk, A. P. & Rotter, S. Shaping the propagation of light in complex media. *Nat. Phys.* **18**, 994–1007 (2022).

22. Popoff, S. M., Lerosey, G., Carminati, R., Fink, M., Boccara, A. C. & Gigan, S. Measuring the Transmission Matrix in Optics: An Approach to the Study and Control of Light Propagation in Disordered Media. *Phys. Rev. Lett.* **104**, 100601 (2010).

23. Marshall, A. W., Olkin, I. & Arnold, B. C. *Inequalities: Theory of Majorization and Its Applications*. (Springer New York, New York, NY, 2011). doi:10.1007/978-0-387-68276-1.

24. Luis, A. Coherence for vectorial waves and majorization. *Opt. Lett.* **41**, 5190 (2016).

25. Streltsov, A., Adesso, G. & Plenio, M. B. *Colloquium*: Quantum coherence as a resource. *Rev. Mod. Phys.* **89**, 041003 (2017).

26. Guo, C., Miller, D. A. B. & Fan, S. Transport Measurements of Majorization Order for Wave Coherence. *Phys. Rev. Lett.* **135**, 053801 (2025).

27. Shen, Y., Harris, N. C., Skirlo, S., Prabhu, M., Baehr-Jones, T., Hochberg, M., Sun, X., Zhao, S., Larochelle, H., Englund, D. & Soljačić, M. Deep learning with coherent nanophotonic circuits. *Nat. Photonics* **11**, 441–446 (2017).

28. Lin, X., Rivenson, Y., Yardimci, N. T., Veli, M., Luo, Y., Jarrahi, M. & Ozcan, A. All-optical machine learning using diffractive deep neural networks. *Science* **361**, 1004–1008 (2018).

29. Hamerly, R., Bernstein, L., Sludds, A., Soljačić, M. & Englund, D. Large-Scale Optical Neural Networks Based on Photoelectric Multiplication. *Phys. Rev. X* **9**, 021032 (2019).

30. Feldmann, J., Youngblood, N., Karpov, M., Gehring, H., Li, X., Stappers, M., Le Gallo, M., Fu, X., Lukashchuk, A., Raja, A. S., Liu, J., Wright, C. D., Sebastian, A., Kippenberg, T. J., Pernice, W. H. P. & Bhaskaran, H. Parallel convolutional processing using an integrated photonic tensor core. *Nature* **589**, 52–58 (2021).

31. McMahon, P. L. The physics of optical computing. *Nat. Rev. Phys.* **5**, 717–734 (2023).

32. Momeni, A., Rahmani, B., Malléjac, M., Del Hougne, P. & Fleury, R. Backpropagation-free training of deep physical neural networks. *Science* **382**, 1297–1303 (2023).

33. Vaswani, A., Shazeer, N., Parmar, N., Uszkoreit, J., Jones, L., Gomez, A. N., Kaiser, L. & Polosukhin, I. Attention is All you Need. in *Advances in Neural Information Processing Systems* vol. 30 (Curran Associates, Inc., 2017).

34. Kim, J.-H., Jun, J. & Zhang, B.-T. Bilinear attention networks. in *Proceedings of the 32nd International Conference on Neural Information Processing Systems* 1571–1581 (Curran Associates Inc., Red Hook, NY, USA, 2018).

35. Vellekoop, I. M. & Mosk, A. P. Focusing coherent light through opaque strongly scattering media. *Opt. Lett.* **32**, 2309 (2007).

36. Guo, C. & Fan, S. Majorization Theory for Unitary Control of Optical Absorption and Emission. *Phys. Rev. Lett.* **130**, 146202 (2023).

37. Guo, C. & Fan, S. Unitary control of partially coherent waves. I. Absorption. *Phys. Rev. B* **110**, 035430 (2024).

38. Guo, C. & Fan, S. Unitary control of partially coherent waves. II. Transmission or reflection. *Phys. Rev. B* **110**, 035431 (2024).

39. Mandel, L. & Wolf, E. Coherence Properties of Optical Fields. *Rev. Mod. Phys.* **37**, 231–287 (1965).

40. Rafayelyan, M., Dong, J., Tan, Y., Krzakala, F. & Gigan, S. Large-Scale Optical Reservoir Computing for Spatiotemporal Chaotic Systems Prediction. *Phys. Rev. X* **10**, 041037 (2020).

41. Zhang, Z., You, Z. & Chu, D. Fundamentals of phase-only liquid crystal on silicon (LCOS) devices. *Light Sci. Appl.* **3**, e213–e213 (2014).

42. Bolduc, E., Bent, N., Santamato, E., Karimi, E. & Boyd, R. W. Exact solution to simultaneous intensity and phase encryption with a single phase-only hologram. *Opt. Lett.* **38**, 3546–3549 (2013).

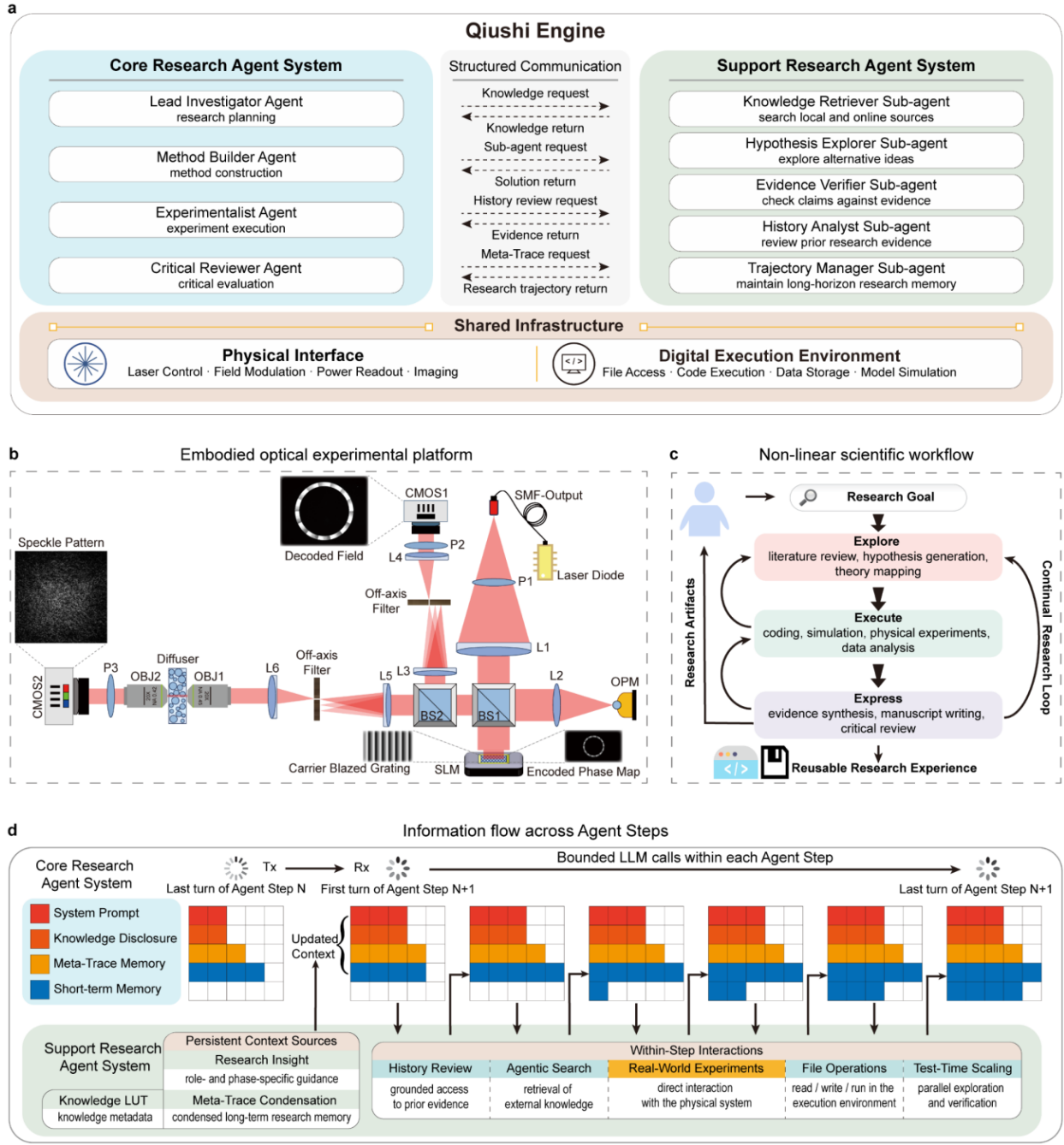

**Figure 1. Qiushi Engine architecture. a**, Dual-layer multi-agent architecture. Qiushi Engine consists of a core research agent system, a support research agent system and a shared infrastructure layer. The core system includes four role-specialized agents: Lead Investigator, Method Builder, Experimentalist and Critical Reviewer. The support system includes modules for history review, knowledge retrieval, hypothesis exploration, trajectory management and evidence verification. The two systems are connected through structured requests and curated returns for history review, knowledge retrieval, hypothesis exploration, trajectory management and evidence verification. The shared infrastructure layer contains a physical interface for hardware control and a digital execution environment for files, code, data, simulation and report generation. **b**, Embodied optical experimental platform. The physical interface connects Qiushi Engine to a benchtop free-space optical system through standardized program modules, including a laser, optical power meter (OPM), spatial light modulator (SLM), reconstructed-field camera (CMOS1) and scattering-output camera (CMOS2). The optical path includes field modulation, beam splitting, Fourier filtering, scattering, optical power monitoring and camera-based detection. **c**, Non-linear scientific workflow. Qiushi Engine operates across three research phases: Explore, Execute and Express. Explore includes literature analysis, hypothesis generation, theory mapping and observable design. Execute includes code implementation, simulation, physical measurement and data analysis. Express includes figure generation, manuscript construction, evidence synthesis and critical review. Agent roles and research phases are shown as decoupled components. **d**, Information flow across Agent Steps. At the boundary between Agent Steps, the system updates the next agent's context through the system prompt, short-term memory, condensed Meta-Trace memory and progressively disclosed knowledge and skills. Within each step, the acting agent interacts with support sub-agents, digital tools and the physical platform. Each step ends with a new Meta-Trace record containing the current state, evidence, artefacts, limitations and next handoff.

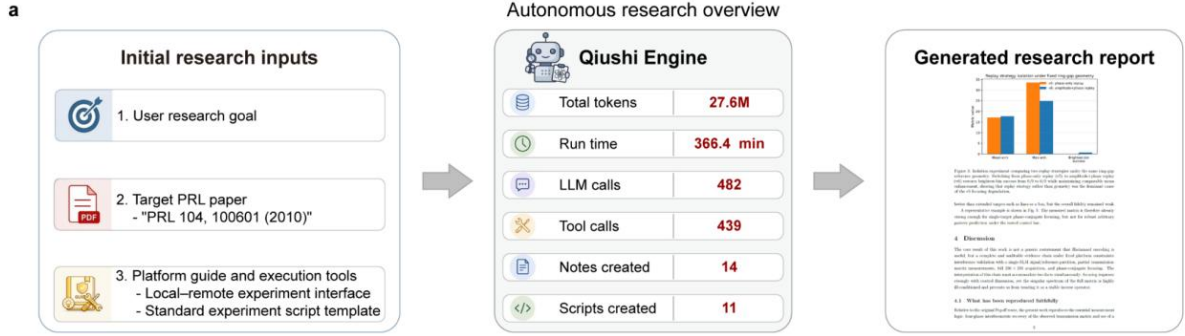

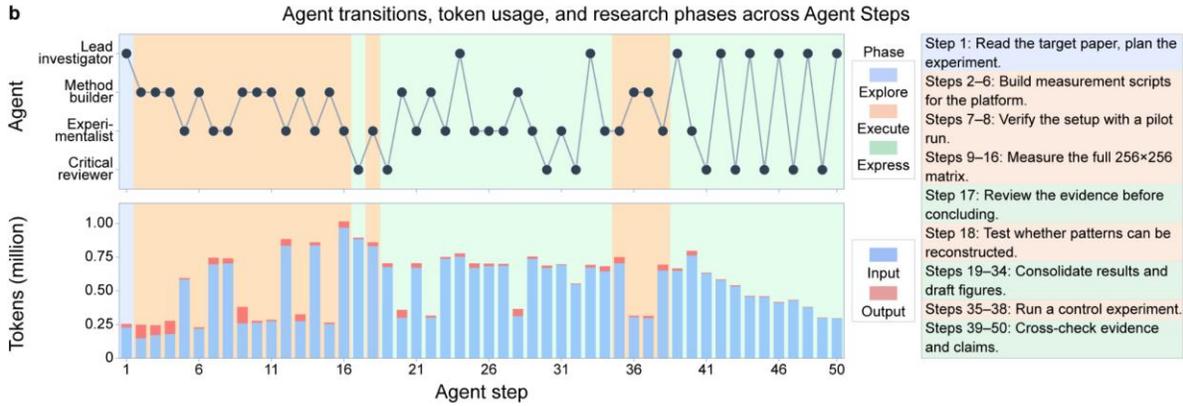

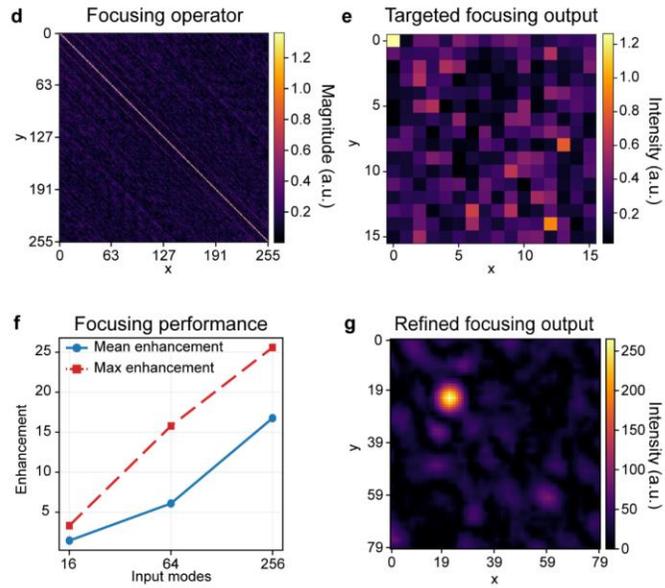

**Figure 2. Autonomous reproduction of a high-impact transmission-matrix experiment on a non-original optical platform. a**, End-to-end research trajectory. Starting from a minimal prompt, the target paper and a basic description of the optical platform, Qiushi Engine develops the reproduction from paper interpretation to experimental execution, data analysis and report generation. **b**, Agent trajectory across 50

Agent Steps. The panels show research phases, agent handoffs and token usage during the reproduction study. **c**, Representative transition between Agent Steps 17 and 18. After obtaining the main transmission-matrix focusing result, the system reviews whether the evidence also supports stronger image- or pattern-reconstruction claims, and returns to experiment for a targeted follow-up test. **d**, Measured focusing operator. The focusing operator constructed from the measured 256 × 256 transmission matrix shows a dominant diagonal structure consistent with mode-selective focusing. **e**, Initial focusing result. Targeted phase-conjugate excitation produces a focused output in a 16 × 16 binned output-mode map. **f**, Mode-number scaling. Focusing enhancement increases as the number of controlled input modes is increased. **g**, Refined focusing result. A follow-up experiment using an optimized reference-field geometry produces an 80 × 80 output-pixel focusing map.

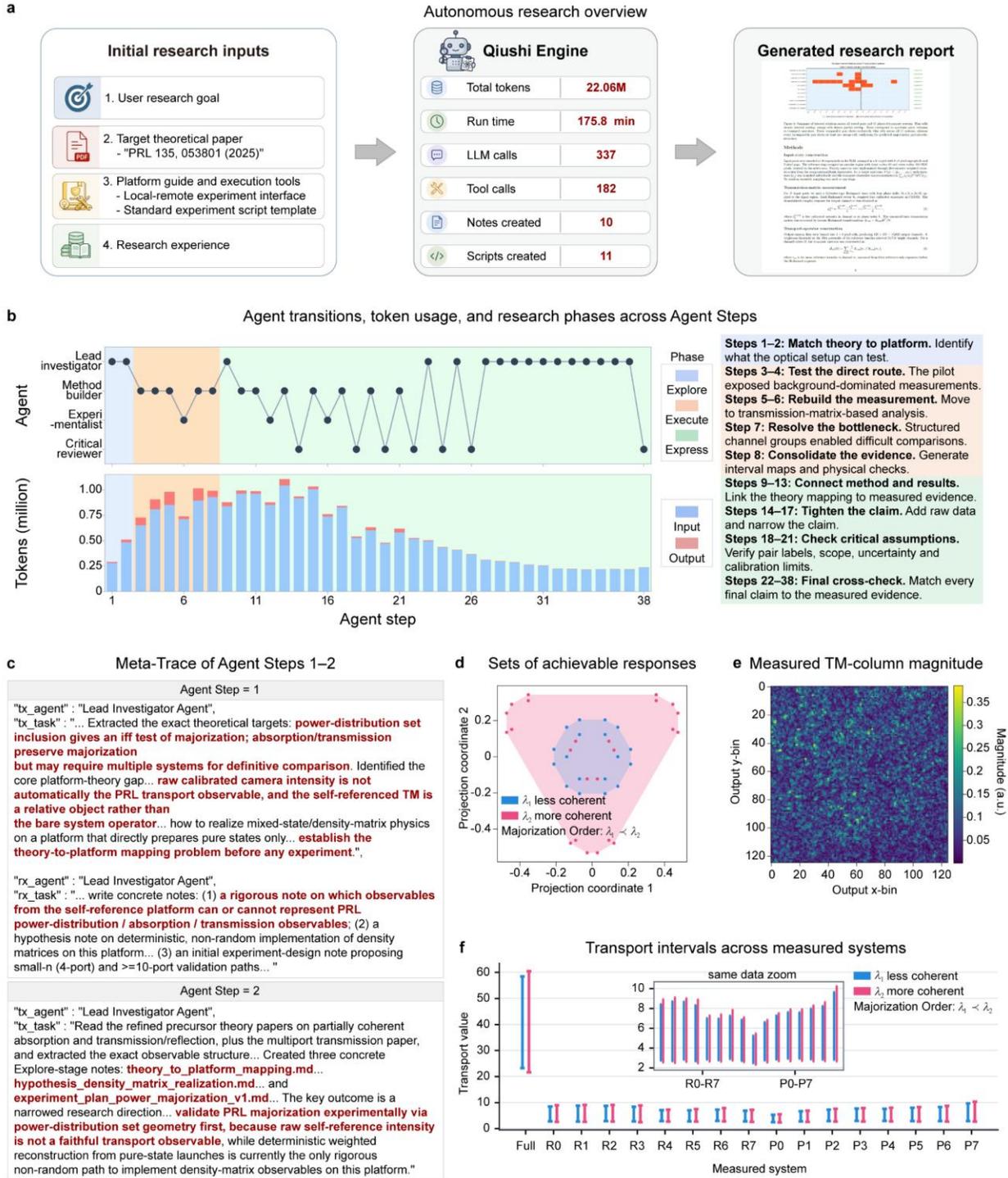

**Figure 3. Autonomous experimental validation of an abstract coherence-order theory on a real optical platform. a**, End-to-end research trajectory. Starting from a minimal prompt, the target theory paper and a basic platform description, Qiushi Engine develops the study from theoretical interpretation to

platform-specific experimental design, measurement, data analysis and report generation. **b**, Agent trajectory across 38 Agent Steps. The panels show agent transitions, research phases and token usage during the theory-to-experiment validation study. **c**, Early theory-to-experiment handoff. The first Agent Step formulates a validation plan that maps the majorization-order transport prediction to measurable response intervals; the following Agent Step converts this plan into concrete measurement and analysis operations on the optical platform. **d**, Predicted response sets. The theoretically predicted achievable-response sets for a representative comparable pair are shown in projected coordinates. **e**, Measured transmission response. The measured transmission-matrix-column magnitude from the final 16-port experiment is displayed over output bins. **f**, Measured transport intervals. The vertical bars denote response intervals computed from transport operators constructed from measured transmission matrices. Across the measured systems, the interval associated with the less coherent spectrum remains nested within that of the more coherent spectrum, consistent with the majorization-order prediction.

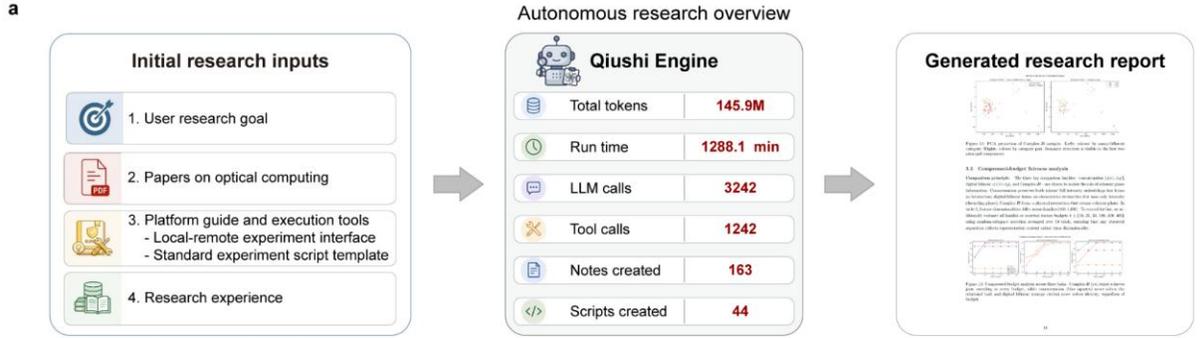

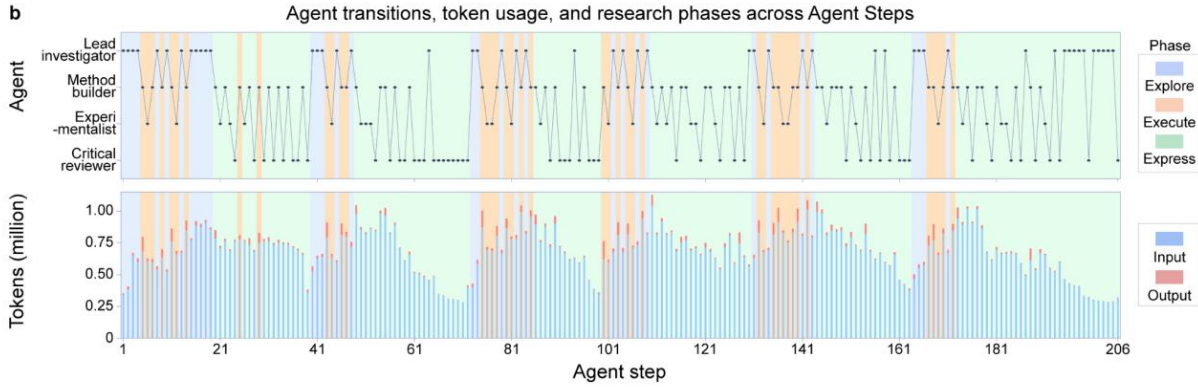

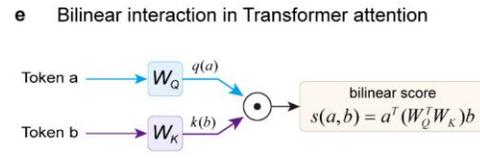

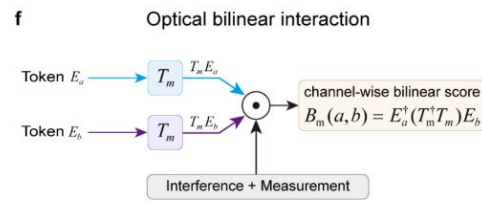

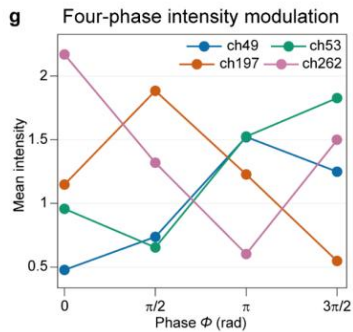

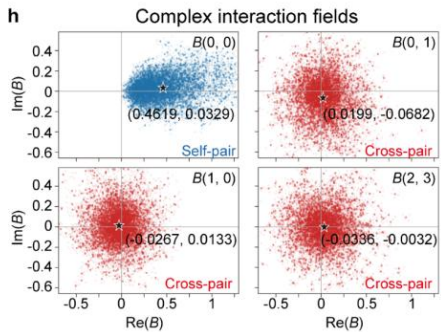

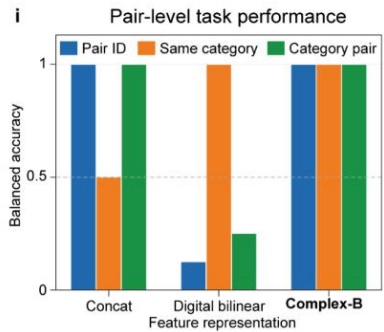

**Figure 4. Autonomous discovery of an optical bilinear interaction for pairwise computation. a**, Overview of the open-ended study. **b**, Agent trajectory across 206 Agent Steps. The panels show agent transitions, research phases and token usage during the open-ended discovery study. **c**, Research directions and refinement. Qiushi Engine first develops four candidate directions: deterministic physical token embedding in complex scattering media, bilinear interaction engines for compact optical pairwise computation, family-controlled physical interaction geometry, and scale-dependent specialization in family-programmable scattering engines. The second direction is then selected for further refinement. **d**, Origin of the bilinear-interaction idea. The Meta-Trace at Agent Step 39 records the transition from single-token scattering embeddings to two-input optical interaction, based on coherent superposition, scattering-induced mixing and square-law detection. **e**, Bilinear compatibility in Transformer attention. Two tokens are projected by learned maps $W_Q$ and $W_K$ into query and key vectors, and their compatibility is computed as a scalar bilinear score. **f**, Channel-wise optical bilinear interaction. Two token-dependent optical fields are coherently superposed, scattered and measured; interferometric demodulation isolates one complex bilinear response at each detector channel. **g**, Four-phase demodulation. Representative detector channels show intensity modulation over four controlled relative phases. **h**, Complex-B distributions. Extracted Complex-B responses for representative ordered input pairs are plotted in the complex plane. **i**, Pair-level classification. Complex-B is compared with token concatenation and an intensity-only digital bilinear baseline under matched linear evaluation.